\documentclass[a4paper,conference]{IEEEtran}

\usepackage{multirow}

\usepackage{cite}

\ifCLASSINFOpdf
  \usepackage[pdftex]{graphicx}
  \graphicspath{{../pdf/}{../jpeg/}}
  \DeclareGraphicsExtensions{.pdf,.jpeg,.png}
\else
  \usepackage[dvips]{graphicx}
  \graphicspath{{../eps/}}
  \DeclareGraphicsExtensions{.eps}
\fi

\usepackage{graphicx}

\usepackage{amsmath}
\usepackage{algorithm}
\usepackage{algpseudocode}

\usepackage{array}

\ifCLASSOPTIONcompsoc
  \usepackage[caption=false,font=normalsize,labelfont=sf,textfont=sf]{subfig}
\else
  \usepackage[caption=false,font=footnotesize]{subfig}
\fi

\usepackage{fixltx2e}

\usepackage{dblfloatfix}

\usepackage{url}
\begin{document}
%
\title{Dynamic Ensemble Selection Using Fuzzy Hyperboxes}

\author{\IEEEauthorblockN{Reza Davtalab}
\IEEEauthorblockA{LIVIA, École de technologie \\supérieure
ÉTS, Montreal, Quebec\\
Email: re.davtalab@gmail.com}
\and
\IEEEauthorblockN{Rafael M.O. Cruz}
\IEEEauthorblockA{LIVIA, École de technologie\\ supérieure
ÉTS, Montreal, Quebec\\
Email: rafael.menelau-cruz@etsmtl.ca}
\and
\IEEEauthorblockN{Robert Sabourin}
\IEEEauthorblockA{LIVIA, École de technologie\\ supérieure 
ÉTS, Montreal, Quebec\\
Email: robert.sabourin@etsmtl.ca}}


%



\maketitle
\setlength{\textfloatsep}{5pt}
\setlength{\abovecaptionskip}{2pt plus 3pt minus 4pt} 
\begin{abstract}
Most dynamic ensemble selection (DES) methods utilize the K-Nearest Neighbors (KNN) algorithm to estimate the competence of classifiers in a small region surrounding the query sample. However, KNN is very sensitive to the local distribution of the data. Moreover, it also has a high computational cost as it requires storing the whole data in memory and performing multiple distance calculations during inference. Hence, the dependency on the KNN algorithm ends up limiting the use of DES techniques for large-scale problems. This paper presents a new DES framework based on fuzzy hyperboxes called FH-DES. Each hyperbox can represent a group of samples using only two data points (Min and Max corners). Thus, the hyperbox-based system will have less computational complexity than other dynamic selection methods. In addition, despite the KNN-based approaches, the fuzzy hyperbox is not sensitive to the local data distribution. Therefore, the local distribution of the samples does not affect the system's performance. Furthermore, in this research, for the first time, misclassified samples are used to estimate the competence of the classifiers, which has not been observed in previous fusion approaches. Experimental results demonstrate that the proposed method has high classification accuracy while having a lower complexity when compared with the state-of-the-art dynamic selection methods. The implemented code is available at https://github.com/redavtalab/FH-DES\_IJCNN.git. 
\end{abstract}


%
\IEEEpeerreviewmaketitle

\section{Introduction}

	Multiple Classifier Systems (MCS) are a good solution for the complex and vast amounts of data that we face today \cite{cruz_dynamic_2018,zyblewski2021preprocessed}. Different types of MCS have been introduced, but many researchers concluded that Dynamic Selection (DS) could be a better choice for the combination of classifiers \cite{britto2014dynamic,cruz_dynamic_2018}. In DS approaches, each given query sample is labeled by an ensemble of base classifiers which are usually selected with regards to their local competence.\par
	Estimation of competence level is a key issue in DS approaches. In this stage, Dynamic Selection Data (DSEL) is used to evaluate the competence level of classifiers. For this purpose, the efficiency of the classifiers in a small region surrounding the query instance on DSEL data is considered as an estimation of the local competence of classifiers \cite{kuncheva2014combining,britto2014dynamic,cruz_dynamic_2018}. This region is called Region of Competence (RoC) and in most of DS approaches, this region is defined either by the K-Nearest Neighbor (KNN) technique applied in the feature space \cite{cruz_meta-des__2015,fernandez-delgado_we_nodate_2014,xiao_ensemble_2016,krawczyk_dynamic_2018,cruz_dynamic_2018,elmi_2020_fuzzyHesitate},  clustering \cite{kuncheva_clustering-and-selectionmodel_nodate,lin_libd3c_2014}, potential functions \cite{woloszynski_new_2009,woloszynski_measure_2012} or the KNN applied in the decision space \cite{giacinto_dynamic_2001,cavalin_logid_2012,batista_dynamic_2012,nguyen_ensemble_2020}.\par
	
	\begin{figure}
		\centering
		\includegraphics[width=1\linewidth]{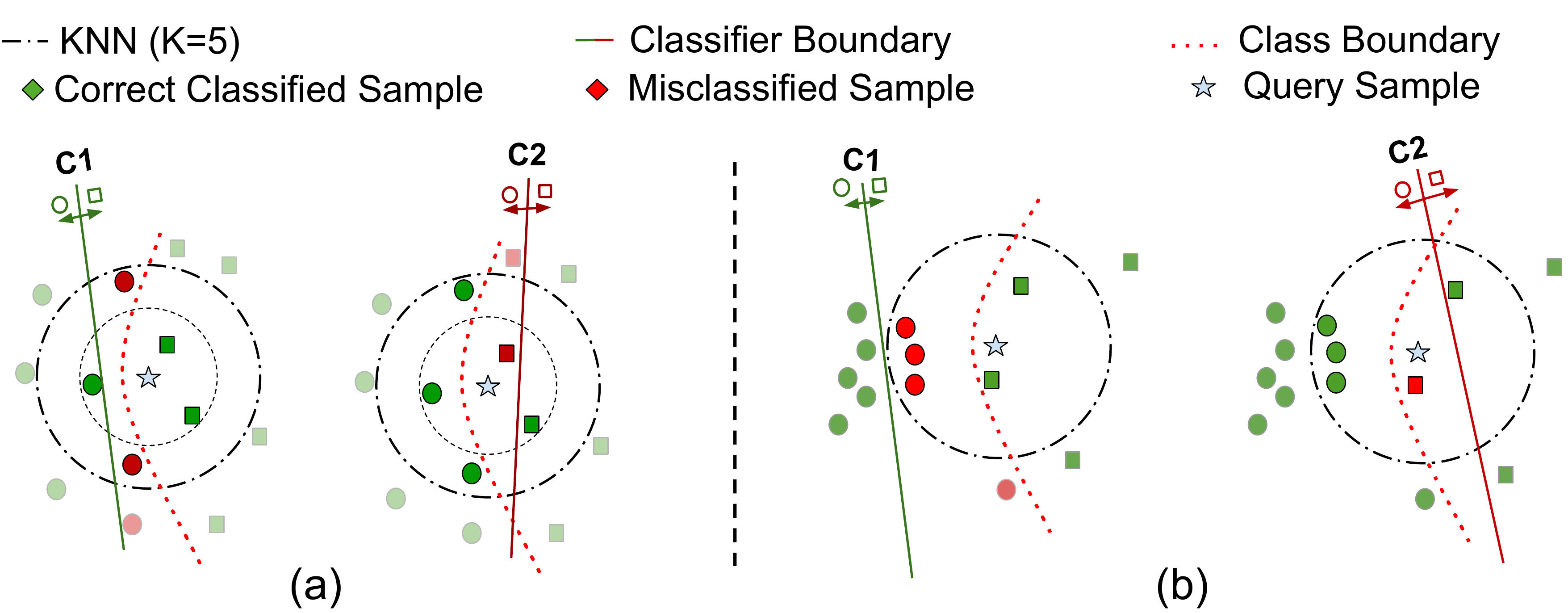}
		\caption{Problems of KNN-based DS approaches. (a) K-value problem, and (b) High sensitivity to local distribution of data. In both cases, just C1 could correctly classify the query sample while KNN (K=5) selects C2 as a competent classifier. }
		\label{fig:knn_problems}
	\end{figure} 
	
    KNN-based approaches are more popular; however, they suffer from high complexity in the generalization phase. In this stage, for each query sample, its k nearest neighbors must be found. This means that the distance between the given data point and all samples must be calculated, which endures the system's huge calculation complexity. Clustering-based approaches reduce this complexity by adopting more coarse grained regions of competence (clusters). They require only require calculating distances to each cluster centroid and then selecting the most competent classifiers according to the nearest cluster. However, the reduction in complexity comes with a significant loss in accuracy compared with KNN-based approaches \cite{cruz_dynamic_2018}.
    
    Moreover, the selected K-value may not work correctly in all regions (K-value problem), even if it has been optimized using an optimization process. As shown in Figure \ref{fig:knn_problems}-a, K=5 results in selecting the wrong classifier (C2) while K=3 can select the correct one. Techniques based on potential function aims to solve this problem by not having a K-value and considering all data points during the competence estimation. In this case, it considers a potential function that gives higher weights to the samples closer to the query while decreasing as the distance increases~\cite{cruz_dynamic_2018}. However, its computational complexity is even higher compared to KNN based approaches as it not only suffers from the high computational cost of calculating the distance between all samples in memory, but it also needs to aggregate the information of the whole set with the application of the potential function.
	
	Additionally, KNN works based on the Euclidean distance that has great sensitivity to the local distribution of data. Hence, a high degree of overlap in the data may lead to a wrong decision (Figure \ref{fig:knn_problems}-b). Finally, KNN just considers the samples of RoC, which contain a limited amount of information. Thus, DS techniques can end up limited to the main problems of the KNN technique and new ways of estimating classifier's competence are needed in order to achieve better classification results while reducing computational cost.

\begin{figure}
		\centering
		\includegraphics[width=1\linewidth]{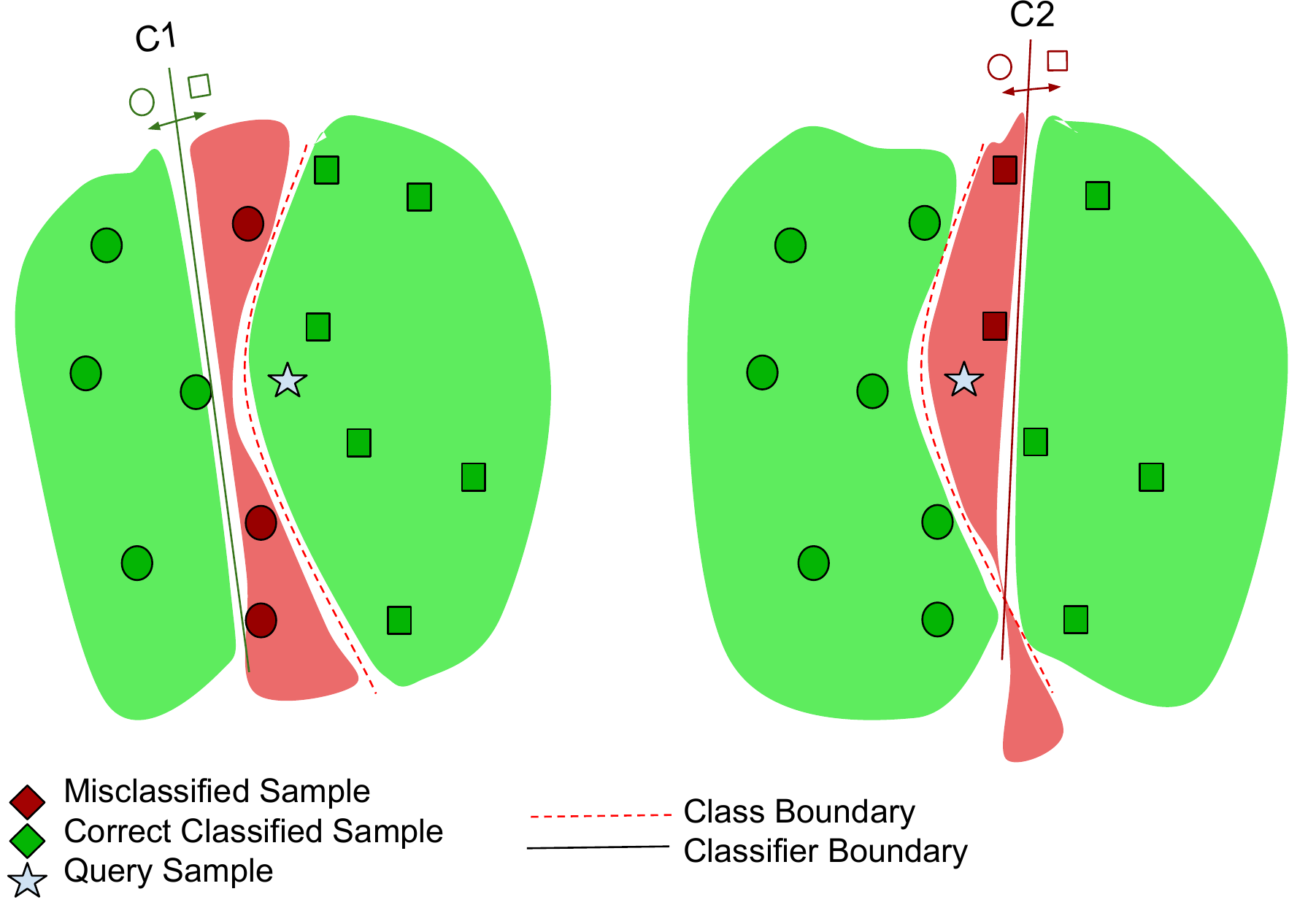}
		\caption{Competence and incompetence areas of classifiers in Figure \ref{fig:knn_problems}-a.}
		\label{fig:example0-Ideal}
	\end{figure}

Intuitively, defining and storing each classifier's competence and incompetence areas could increase the labeling speed in the generalization phase. Falling the query sample $\mathbf{x_q}$ into the competence region of $c_i$ means that this classifier is competent to classify $\mathbf{x_q}$.
Figure \ref{fig:example0-Ideal} illustrates the initial idea of this approach to solve the example of Figure \ref{fig:knn_problems}-a. 
	In this example, the competence of C1 is estimated higher than C2, because the query sample has fallen in the green area of C1. However, defining the domain of such areas is not easy and imposes a large computational complexity on the system, unless some simpler structures are used to represent these areas. In a two-dimensional feature space, we can represent these areas using rectangles. Each rectangle could be defined by only two points. Therefore, its computational complexity will not be high if there are an acceptable number of rectangles to represent all training samples. \par 
	
    \textit{Hyperbox} is a virtual concept that works like these rectangles; however, it is capable of working in high-dimensional spaces. Each hyperbox covers the interior space and a small part of its vicinity. As we move away from the hyperbox, its coverage decreases fuzzily according to a membership function. That is why it is called \textit{fuzzy hyperbox} \cite{simpson1992Classification}. The fuzzy aspect of the hyperbox gives us valuable information outside of the hyperbox, and we can estimate how far the query sample is from the competence or incompetence area of the base classifier. Thus, we can estimate the competence of classifiers even if the query sample falls outside of all hyperboxes.
	We will discuss hyperboxes in detail in the section \ref{sec:FuzzyHyperboxeBackground}. \par
\begin{figure}
		\centering
		\includegraphics[width=1\linewidth]{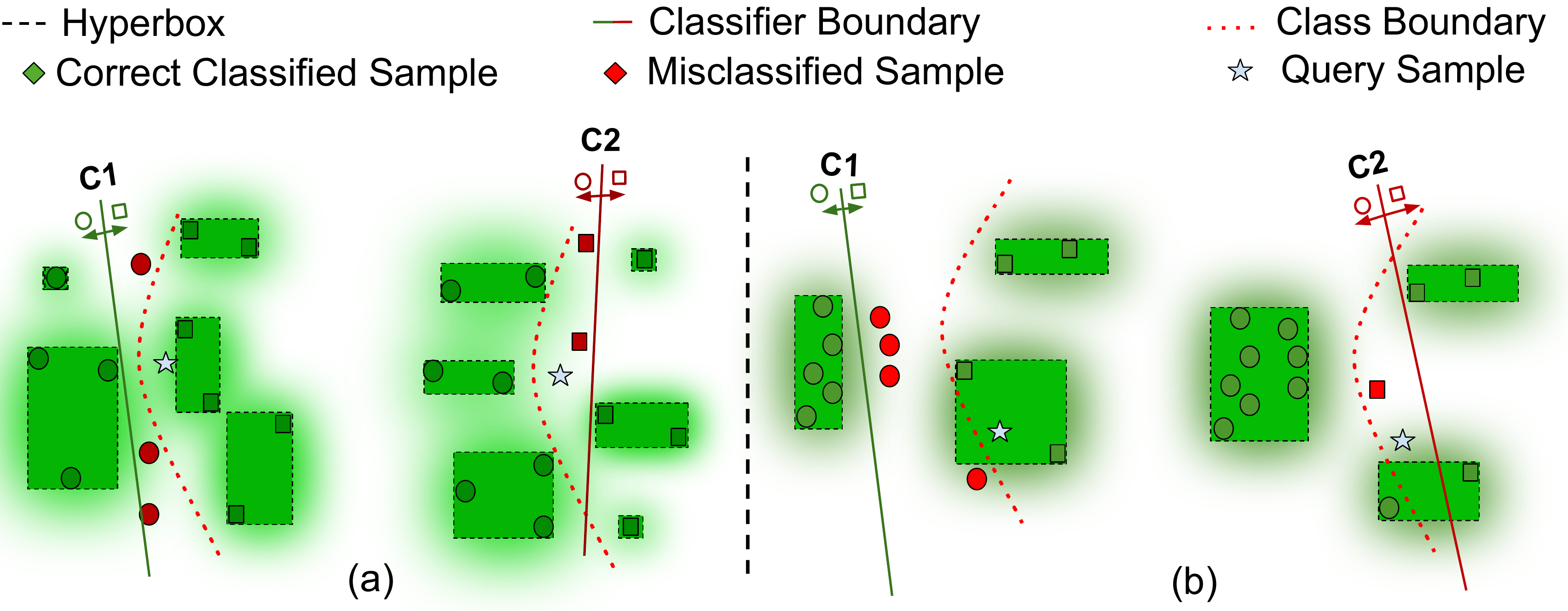}
		\caption{Solving the illustrated problems of KNN in Figure \ref{fig:knn_problems} using Fuzzy Hyperboxes based on correct classified samples that represent competence areas}
		\label{fig:Hyperboxes_solutions}
	\end{figure}

 
	In Figure \ref{fig:Hyperboxes_solutions}, fuzzy hyperboxes are used to represent the competence regions in the examples of Figures \ref{fig:knn_problems}. As illustrated in Figure \ref{fig:Hyperboxes_solutions}-a, the query sample is located outside of all hyperboxes. However, it is located close to the hyperbox of classifier C1 (inside its green area of C1). Therefore, this classifier is considered to be more competent than classifier C2.\par	

	In summary, this research aims to answer the following research questions: (1) Can the use of fuzzy hyperboxes lead to more accurate dynamic selection approaches? (2) Do the miss-classified samples have enough information to estimate the competence of classifiers? (3) Will the use of hyperboxes lead to reduced computational complexity compared to current DS techniques?  
	
	The rest of the paper is organized as follows: In Section \ref{sec:FuzzyHyperboxeBackground}, the background of Fuzzy Hyperboxes is reviewed. The proposed method is discussed in Section \ref{sec:ProposedFramework}. Finally, the experimental results and conclusion are discussed in Section \ref{sec:ExperResults} and Section \ref{sec:conclusion}, respectively.

\section{Fuzzy Hyperbox}
    \label{sec:FuzzyHyperboxeBackground}

	Hyperbox was introduced by Simpson in 1992 as a building block for Fuzzy Min-Max Neural Networks (FMM) \cite{simpson1992Classification,simpson_fuzzy_1993}. Hyperbox is defined by its two corners named \textit{Min ($\mathbf{v}$)} and \textit{Max ($\mathbf{w}$)} corners.The size and location of the hyperboxes are easily adjustable by changing these two corners. Hyperbox-based learning systems have some features that make them promising tools in machine learning applications: the ability to make soft and hard decisions, scalability, online adaptation, and the ability to model granular data \cite{khuat_hyperbox_2019}.
	
	\subsection{Learning process}
	\label{subsec:learningProcessHyperbox}
\begin{figure}
		\centering
		\includegraphics[width=0.85\linewidth]{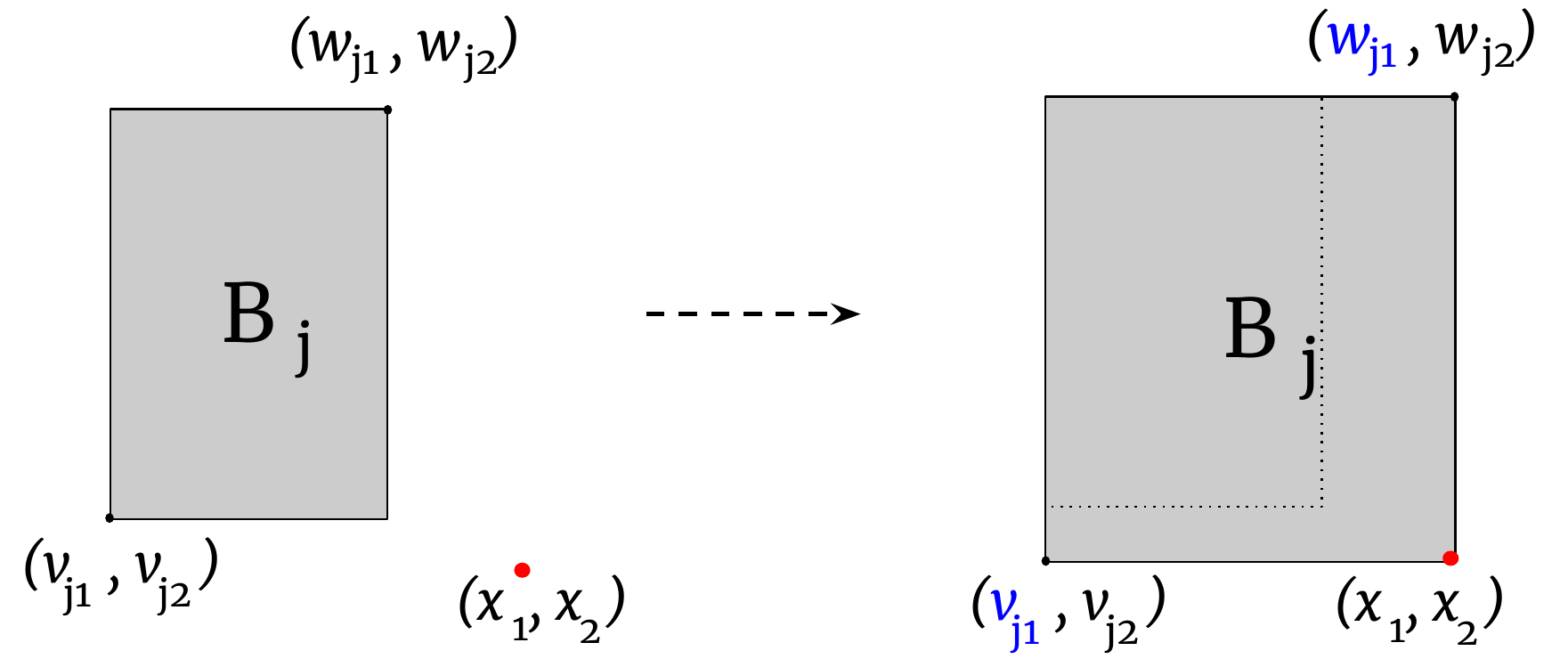}
		\caption{Expansion of hyperbox $B_j$ to involve sample $\mathbf{x} (x_1,x_2)$.}
		\label{fig:expansion}
	\end{figure}

	The learning process of hyperboxes is a single-pass process in which hyperboxes are formed regarding to learning data to cover the needed regions. During this process, for each learning instance $\mathbf{x}$, a hyperbox must be found that contains $\mathbf{x}$ or is expandable enough to contain this sample. Figure \ref{fig:expansion} shows how the hyperbox $B_j$ is expanded to involve the sample $\mathbf{x}$. In this example, $v_{j2}$ and $w_{j1}$ are changed to expand the hyperbox. 
	
	The maximum size of hyperboxes is limited by the user-defined parameter $\theta$ during the learning process. At the end of this process, if no expandable hyperbox is found, a new hyperbox is created. 
	\par  
	
	Each hyperbox is defined by the following equation.
	\begin{equation}
	B_j=\{\mathbf{v_j},\mathbf{w_j},b_j(\mathbf{x})\}  \:  \forall\, \mathbf{x} \in I^n 
	\label{eq:Bj}
	\end{equation}
	
	In this equation, $\mathbf{x}=\{x_1,x_2,. . .,x_n\}$ is a single data point. $\mathbf{w_j}=\{w_{j1},w_{j2},. . .,w_{jn} \}$ and $\mathbf{v_j}=\{v_{j1},v_{j2},. . .,v_{jn} \}$ are min and max corners of the hyperbox, respectively. $b_j$ is the membership function of the hyperbox $B_j$. Also, $I^n$ is $n$ dimensional feature space. 
	
	\subsection{Membership Function}
	The membership function of hyperbox is a crucial part of the fuzzy Min-Max neural network technique. It is utilized to quantify the membership grade of an arbitrary instance to the hyperbox $B_j$ (between 0 and 1). The membership function of a hyperbox is usually defined so that the degree of membership inside the hyperbox $B_j$ is equal to one and decreases when the feature point moves away from the hyperbox. \par
	Many membership functions were proposed for hyperboxes. However, the membership function introduced by Gabrys and Bargiela \cite{gabrys_general_2000} is the most popular membership function among the fuzzy hyperbox's applications \cite{khuat2020hyperbox}. It has a simple structure (only the min and max points), and the membership value monotonically decreases by increasing each side of the hyperbox. This function is defined as follows:
	
	\begin{eqnarray} \nonumber
	b_j(\mathbf{x})\!&=&\!min_{i=1..n}(min([1-f(x_{i} - w_{ij},\gamma_i )],\\ 
	&& [1-f(v_{ij} - x_{i},\gamma_i )]  )).
	\label{eq:FM_gfmm} 
	\end{eqnarray}
	Where,
	\begin{equation}
	f(r,\gamma)= \left\{
	\begin{array}{rl}
	1 &\mbox{if $r\gamma > 1$}\\ 
	r\gamma &\mbox{if $0\leq r\gamma\leq 1$} \\ 
	0 &\mbox{if  $r\gamma<0$}
	\end{array} \right.
	\end{equation}

\begin{figure}
		\centering
		\includegraphics[width=0.8\linewidth]{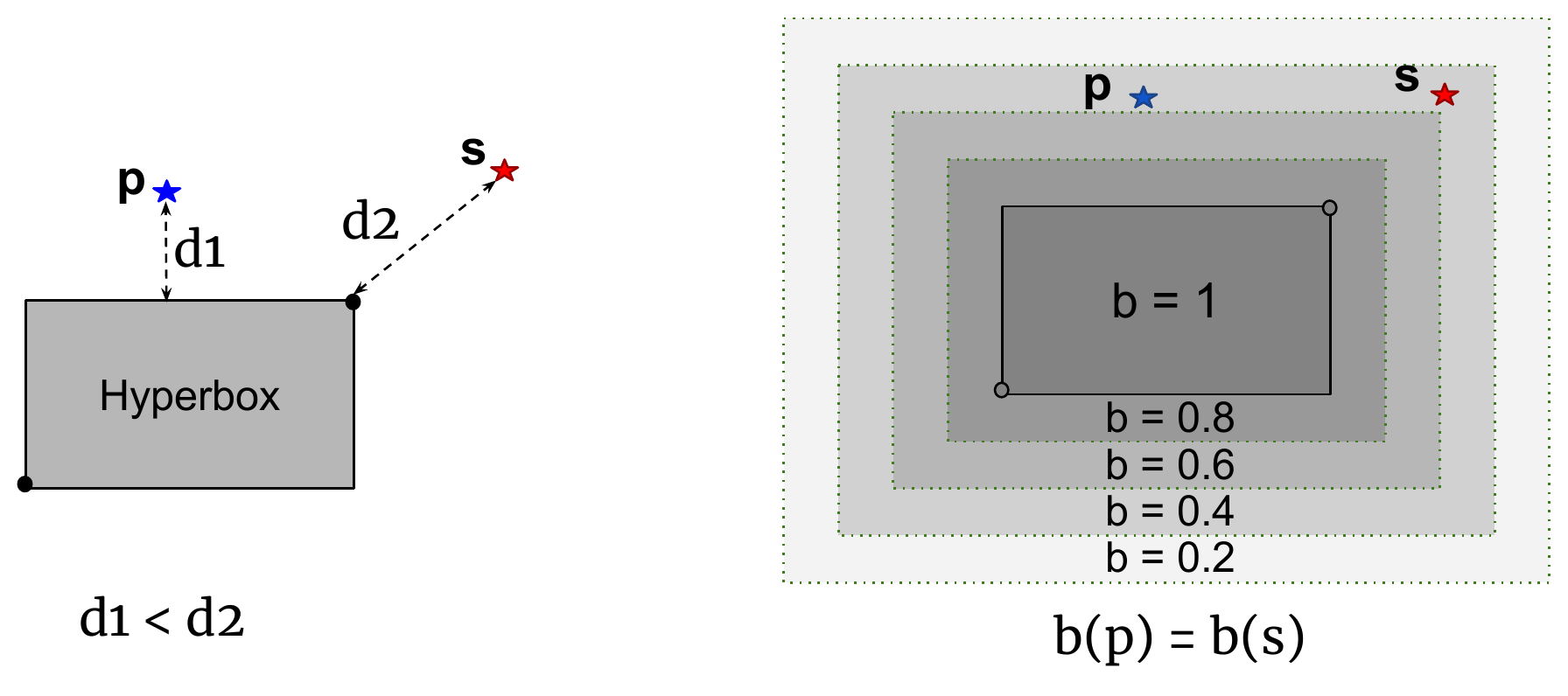}
		\caption{Membership function proposed by Gabrys and Bargiela and its corners problem}
		\label{fig:GFMM_corners problem}
	\end{figure}

    Where $a_{i}$ is $i_{th}$ dimension of a sample $\mathbf{x}$, and $\gamma$ is the sensitivity parameter that regulates the rate with which the membership values decrease out of the hyperbox. 
	However, this membership function has sharp corners, which assigns a higher membership to further samples in some cases. Membership levels around the hyperbox and the mentioned problem of this membership function are shown in Figure \ref{fig:GFMM_corners problem}. \par
	
	As can be observed, this function assigns the same membership value to the points P and S, while P is closer to the hyperbox than S. 

\section{Proposed Framework (FH-DES)}
\label{sec:ProposedFramework}
	Here, a novel DS framework based on fuzzy hyperboxes is introduced, called Fuzzy Hyperbox-based Dynamic Ensemble Selection (FH-DES). In this approach, the competence or incompetence areas of classifiers (Figure \ref{fig:example0-Ideal}) are defined by fuzzy hyperboxes.

	When hyperboxes are built based on correctly classified examples (Figure \ref{fig:Hyperboxes_solutions}), they represent regions where the classifiers work well or competence areas. This approach will be called FH-DES-C in the rest of the paper. In contrast, when hyperboxes are built with misclassified samples, the approach is called FH-DES-M, in which hyperboxes represent areas of incompetence. In this case, the classifier whose hyperboxes have a lower membership degree will be farther from query sample $\mathbf{x_q}$, so it will be more competent to classify the query sample.\par 
	Therefore, the competence of the classifier $c_i$ to classify the given sample $\mathbf{x_q}$ is estimated according to the membership function of hyperboxes that belong to $c_i$. \par 
	As mentioned in section \ref{sec:FuzzyHyperboxeBackground}, Gabrys's membership function \cite{gabrys_general_2000} has some problems in the corners of membership levels. To fix these problems, we introduce a new membership function with smoother borders (SBM): 
	
	\begin{equation}
	b_j(\mathbf{x_q}) = (|| ReLU(|\mathbf{o_j} - \mathbf{x_q}| - (\mathbf{w_j} - \mathbf{v_j})/2 )||_2)^2 
	\label{eq:newMF}
	\end{equation}
	
\begin{figure}
		\centering
		\includegraphics[width=0.4\linewidth]{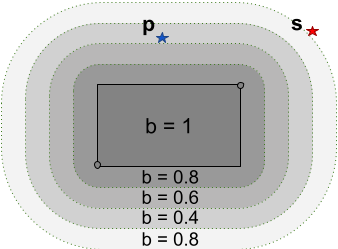}
		\caption{The proposed Smooth-Border membership function (SBM) for FH-DES framework}
		\label{fig:newfm}
	\end{figure}

	Where $\mathbf{o_j}$ is the center of hyperbox $B_j$, $\mathbf{v_j}$ and $\mathbf{w_j}$ are min and max corners respectively, $||.||_2$ indicates 2-norms, and  ${ReLU}(\cdot)$ is the Rectified Linear Unit (ReLU) function as below:
	\begin{equation}
	ReLU(a) = max (0,a) 
	\label{eq:relu}
	\end{equation}
	In Figure \ref{fig:newfm}, the membership levels of SBM is illustrated. The smooth borders of this function help us to solve the mentioned problem in Figure \ref{fig:GFMM_corners problem}.  
	
	In this method, all necessary calculations to define competence (or incompetence) areas are performed during the training phase, and only membership values are calculated to label the query sample $\mathbf{x_q}$ during the generalization phase. Therefore, the proposed framework is expected to have less complexity than KNN-based approaches. 
	In addition, the computational complexity of FH-DES-M should be less than FH-DES-C. Because the number of misclassified samples is usually less than the correctly classified samples. \par
\subsection{Training phase}
During the training phase, after generating the pool of classifiers, all the needed hyperboxes are formed according to the performance of the base classifiers (on DSEL data). It consists of the expansion process of FMM \cite{simpson_fuzzy_1993}. \par 
In particular, suppose we want to use the FH-DES-M approach, and $Mset_i$ contains misclassified samples of classifier $c_i$, all hyperboxes of $c_i$ are built using $Mset_i$ according to the learning process of hyperboxes (Subsection \ref{subsec:learningProcessHyperbox}). The set of hyperboxes, which belongs to the classifiers $c_i$, is called $Hset_i$. The distribution of hyperboxes depends on the order of the samples within $Mset_i$. Consequently, some hyperboxes can overlap in the feature space; However, it does not affect the system's performance.

\begin{figure}
		\centering
		\includegraphics[width=0.85\linewidth]{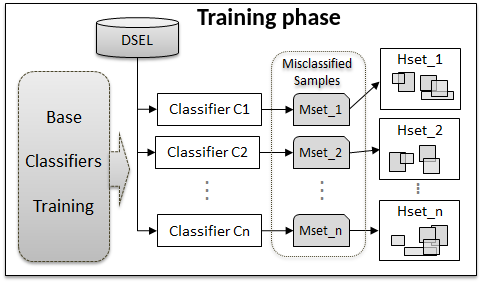}
		\caption{Block Diagram of training phase of FH-DES based on misclassified samples.}
		\label{fig:fhdestrainphase}
	\end{figure}

Figure \ref{fig:fhdestrainphase} represents the training phase of the proposed approach based on misclassified samples. 
	
	As mentioned in Subsection \ref{subsec:learningProcessHyperbox}, the hyperbox creation process for the classifier $c_i$ begins by picking a sample of $Mset_i$ and finding a hyperbox of $Hset_i$ that includes (or can expand to include) the picked sample. If such hyperbox is not found, a new hyperbox is created at the same point and added to $Hset_i$.

\subsection{Generalization phase} 
		
During the generalization phase, for each given query sample $\mathbf{x_q}$, the performance of all classifiers is estimated based on their hyperboxes, and the best ensemble of classifiers is selected. Specifically, for each query sample $\mathbf{x_q}$,
the competence of the classifier $c_i$, represented by $\delta_i$, is calculated as follows:

	\begin{equation}
	\delta_i(\mathbf{x_q}) = (b^{i*} + b^{i+} )/2 
	\label{eq:DeltaC}
	\end{equation}

Here $b^{i*}$ and $b^{i+}$ are the first and second highest membership values among the hyperboxes of $c_i$. It should be noticed that in the correct-classified version (FH-DES-C), the membership value of hyperboxes is related to the competence of the classifier. While in FH-DES-M, the membership values of hyperboxes represent the incompetence of classifiers. Thus, the calculated competence value should be deducted from 1 ($\delta_i \leftarrow 1- \delta_i$) in the FH-DES-M version. \par 
In the next step, the ensemble of the most competent classifiers is selected according to a global threshold $\tau_q$. The threshold is defined by Equation~\ref{eq:threshold}.

	\begin{equation}
	\tau_q = \mu \times \max_{i=1..M} (\delta_i(\mathbf{x_q}))
	\label{eq:threshold}
	\end{equation}
	In this equation, $\mu$ is a predefined parameter between 0 and 1. Therefore, the final ensemble of classifiers ($\phi$) is formed considering threshold $\tau_q$ as below:
	\begin{equation}
	\phi(\mathbf{x_q}) = \{ c_i  |  \delta_i(\mathbf{x_q}) \geq \tau_q  \} 
	\label{eq:selector}
	\end{equation}   
	
	Where $\mu$ is equal to one, only the most competent classifier(s) is selected. On the contrary, when $\mu$ equals zero, all classifiers will be selected. Finally, in the aggregation step, outputs of the selected classifiers are combined with weighted majority voting by associating competence to the classifiers as their weights:
	
	\begin{equation}
	\hat{y} = arg \max_\lambda \sum_{\forall l \  \in \lambda } \delta_i(\mathbf{x_q}) \quad | \quad c_i(\mathbf{x_q})=l , c_i \in \phi(\mathbf{x_q})
	\label{eq:yhat}
	\end{equation}
	
	Where $\lambda$ is the set of unique class labels and $l$ represents label of $\mathbf{x_q}$. The pseudocode of the generalization phase is represented in the algorithm \ref{alg:aggregation}. 

\begin{algorithm}
    \caption{ Labeling process of $\mathbf{x_q}$} \label{alg:aggregation}
    \begin{algorithmic}[1]
        \Require $\omega , \mathbf{x_q}$ 
        \State Calculate the membership value of $\mathbf{x_q}$ for all hyperboxes. 	
		\State Calculate competence of all classifiers by eq (\ref{eq:DeltaC}) 
		\State Select the ensemble of classifiers by eq (\ref{eq:selector})
		\State Aggregate outputs of selected classifiers by eq (\ref{eq:yhat}) 
		\State Label the given sample
	\end{algorithmic}

\end{algorithm}

	In summary, unlike previous DS techniques, in this approach, the strength (competence) or weakness (incompetence) of the classifiers is considered to select the final ensemble of the classifiers, while in current DS approaches, only their strength (competence) is considered as a selection criterion \cite{cruz_dynamic_2018}.\par 
	In addition, the approach solves some problems of the KNN-based approaches, such as the K-value problem, sensitivity to the local distribution, and limited information. Therefore, we expect the proposed method has higher accuracy than KNN-based approaches. In addition, since in the proposed framework, only hyperboxes are utilized in the generalization phase (instead of instances), and the framework contains fewer hyperboxes than samples, it is expected that the proposed approach will be faster than KNN-based DS techniques. This is especially true in the FH-DES-M approach, where hyperboxes are formed based on only misclassified samples.

 	\section{Experimental results}
	\label{sec:ExperResults}
	In this section, the efficiency of the proposed framework is evaluated using 30 datasets and compared with other DS approaches. To have a better comparison, we used a similar experimental protocol that is used in \cite{cruz_dynamic_2018} and \cite{cruz_meta-des__2015}. In this experiment, the pool of classifiers contained 100 perceptrons that were generated using the bagging technique \cite{breiman1996bagging}. This pool was fixed for all techniques to ensure a fair comparison.
	In these experiments, each dataset was randomly divided into 50\% training data, 25\% the dynamic selection dataset (DSEL), and 25\% test data. The division was performed by maintaining the prior probability of each class. To implement different algorithms, DESlib toolkit (version 0.3.5) \cite{cruz_deslib_nodate} was used. Furthermore, each experiment was conducted using 20 replications and the mean of the evaluation criteria has been reported.
	In all experiments, the perceptron of the SciKit-learn library (Version: 1.0.1) is used as a base classiﬁer. Some of the parameters required in these experiments are reported in Table \ref{tbl:protocol}.
	
	\begin{table}[]
		\centering
		\caption{Used parameters and their values in experiments}
		\label{tbl:protocol}
		\small
		\scalebox{0.9}{
		\begin{tabular}{lll}
			\hline
			\textbf{Method} & \textbf{Parameter Name}       & \textbf{Value}        \\ \hline
			\multirow{3}{*}{Perceptron}        & Maximum iteration    & 100          \\  
						& Tolerance            & 10e-3        \\  
						& Alpha                & 0.001        \\ \hline
			
			\multirow{2}{*}{CalibratedClassifierCV} & CV    & prefit       \\ 
			 		& Calibration method   & isotonic     \\ \hline
			
			\multirow{3}{*}{Bagging}      & Number of estimators & 100          \\
					& bootstrap            & True         \\ 
			 	& max- samples         & 1.0          \\ \hline
			
			KNN-based DS  & K  & 7  \\ \hline
			
			\multirow{2}{*}{META-DES}            & K\_p                 & 5            \\ 
			 		& h\_c                 & 80\%         \\ \hline

		\end{tabular}}
		
	\end{table}
	The proposed framework has two main hyperparameters, including Theta ($\theta$), which defines the maximum size of hyperboxes, and Mu ($\mu$) which defines a threshold to select top base classifiers. Both hyper-parameters are set in the range 0 to 1. To tune these hyper-parameters, we use a similar process to that used by Cruz et al.\cite{cruz_meta-des__2015}. Ten different datasets, which have not been used in the comparative study, were used during this process to avoid biased estimation. The tuning experiments were carried out using the same experimental protocol and the optimal values found were $\theta = 0.27$ and $\mu = 0.99$. Therefore, these values were used in the main experiments. \par 
	All simulation details are available in FH-DES's GitHub repository\footnote{https://github.com/redavtalab/FH-DES\_IJCNN.git}.

	\subsection{Datasets}
	\label{ss:datasets}
	
	In our experimental study, 40 different real-world datasets in a wide variety of areas were used. 
	These datasets were selected with different specifications to evaluate different aspects of the proposed approach better. All datasets were collected from OpenML \cite{openml-van2013}, UCI \cite{asuncion2007uci} repositories, and previous DS researches.
	In Table \ref{tbl:datasets} the utilized datasets and their specifications are listed. The first ten datasets in this table were used to tune the proposed approach's hyperparameters. The other 30 datasets were used in our comparative study.

\begin{table}[]
    \centering
	\caption{ Dataset considered in this work and their main features.}
	\label{tbl:datasets}
	\scalebox{.75}{
    \begin{tabular}{llll |  llll}
		\hline 
\# & Database         & Instances & Features    & \# & Database         &  Instances & Features  \\		\hline
1 &             Adult &     690  &       14     &21 &            Heart &     270  &       13    \\  
2 &            Audit2 &     776  &       17     &22 &             ILPD &     583  &       10    \\  
3 &               CTG &    2126  &       21     &23 &       Ionosphere &     351  &       34    \\ 
4 &  Cardiotocography &    2126  &       21     &24 &       Laryngeal1 &     213  &       16    \\  
5 &             Chess &    3196  &       36     &25 &       Laryngeal3 &     353  &       16    \\   
6 &  Credit-screening &     690  &       15     &26 &       Lithuanian &     600  &        2    \\
7 &             Ecoli &     336  &        7     &27 &            Liver &     345  &        6    \\
8 &             Glass &     214  &        9     &28 &     Mammographic &     830  &        5    \\
9 &                P2 &     1000 &        2     &29 &            Monk2 &     432  &        6    \\
10 &      Transfusion &     748  &        4     &30 &          Phoneme &    5404  &        5    \\
\cline{1-4}
11 &            Audit &     771  &       26      &31 &             Pima &     768  &        8  \\ 
12 &           Banana &    1000  &        2      &32 &            Sonar &     208  &       60  \\ 
13 &         Banknote &    1372  &        4      &33 &          Statlog &    1000  &       20  \\
14 &            Blood &     748  &        4      &34 &            Steel &    1941  &       27  \\ 
15 &           Breast &     569  &       30      &35 &          Thyroid &     692  &       16  \\ 
16 &              Car &    1728  &        6      &36 &          Vehicle &     846  &       18  \\ 
17 &         Datauser &     403  &        5      &37 &        Vertebral &     310  &        6  \\ 
18 &           Faults &    1941  &       27      &38 &           Voice3 &     238  &       10  \\ 
19 &           German &    1000  &       24      &39 &          Weaning &     302  &       17  \\ 
20 &         Haberman &     306  &        3      &40 &             Wine &     178  &       13  \\ 

\hline
\end{tabular}}
\end{table}

	\subsection{Performance of the proposed framework in different configurations}
	To determine the best configuration of the proposed framework, we compare their accuracy over the 30 datasets. Thus, we examined the influence of misclassified samples and the smooth corner membership function on the performance of the proposed framework. Therefore, there are four different configurations of the proposed framework. FH-GC and FH-GM approaches are based on the Gabrys membership function, which uses correct-classified and misclassified samples, respectively. Two others utilize the proposed membership function (SBM). This group contains FH-DES-C and FH-DES-M based on correct-classified and misclassified samples, respectively. The classification accuracy of these approaches alongside their standard deviations is reported in Table \ref{tbl:recResults}.
	\begin{table}
	\centering
	\caption{Average accuracy and standard deviation of the proposed method in different configurations}
	 \label{tbl:recResults}
	\scalebox{0.8}{
	    \begin{tabular}{lllll}
\hline
        DataSets &       FH\_GC &       FH\_GM &    FH\_DES-C &    FH\_DES-M \\
\hline
           Audit & 96.81(0.77) &  \textbf{96.94(0.8)} & 96.87(0.74) & 96.87(0.92) \\
          Banana &  89.26(2.5) &  89.4(2.23) & 89.12(2.37) &  \textbf{89.5(1.68) }\\
        Banknote &  99.1(0.57) & \textbf{99.52(0.51)} & 99.13(0.66) &  99.34(0.5) \\
           Blood & \textbf{77.57(2.32)} & 77.22(2.18) & 77.46(1.98) & 76.55(2.64) \\
          Breast & 96.47(1.52) & \textbf{96.82(1.49)} & 96.29(1.49) & 96.61(1.61) \\
             Car & 73.26(1.24) & 72.96(1.09) & \textbf{74.63(1.08)} & 74.11(1.25) \\
Datausermodeling & 87.62(3.66) & 91.19(4.03) & 87.77(3.93) & \textbf{91.29(3.63)} \\
          Faults &  69.31(2.1) & 69.74(2.28) & 70.12(1.84) & \textbf{70.38(2.03)} \\
          German & \textbf{74.94(2.02)} & 74.92(2.28) &   74.8(2.0) & 74.86(2.27) \\
        Haberman & 71.23(3.99) & \textbf{71.56(3.78)} & 71.36(3.88) & \textbf{71.56(3.98)} \\
           Heart & 83.01(3.82) &  82.43(4.2) & 82.79(3.89) &  \textbf{83.9(4.42)} \\
            ILPD &  70.92(3.7) &  70.34(3.7) &   \textbf{71.3(3.1) }& 70.65(2.59) \\
      Ionosphere & 88.75(2.43) & \textbf{88.75(1.52)} & 88.35(2.21) & 88.13(1.37) \\
      Laryngeal1 & 81.85(3.69) & 81.85(3.29) & 81.94(3.25) &  \textbf{82.5(3.49)} \\
      Laryngeal3 &  71.8(3.46) & \textbf{72.13(4.46)} & 71.01(4.15) &  71.8(4.08) \\
      Lithuanian &  89.3(2.25) & \textbf{90.57(2.21)}& 89.63(2.08) &  90.5(2.32) \\
           Liver & 66.72(3.85) & \textbf{69.43(4.88)} & 67.82(4.77) & 69.14(4.34) \\
    Mammographic & 78.73(2.87) & \textbf{79.18(2.82)} &  78.03(3.2) & 78.87(2.73) \\
           Monk2 & 79.03(3.18) & 81.81(3.55) &  86.2(2.85) & \textbf{87.64(3.24)} \\
         Phoneme & 77.73(0.91) & \textbf{78.13(0.89)} & 77.81(0.97) &  78.1(0.91) \\
            Pima & 75.47(1.91) & 75.68(2.41) &  74.9(2.46) & \textbf{76.28(2.72)} \\
           Sonar &  80.29(5.5) & 80.48(5.84) & \textbf{81.35(6.41)} & 79.62(5.42) \\
         Statlog &   75.1(2.6) & 74.94(2.32) & \textbf{75.26(1.91)} &  75.08(2.0) \\
           Steel & 70.64(1.58) & 70.57(1.34) & 70.72(1.67) & \textbf{71.37(1.33)}\\
         Thyroid & 95.92(1.56) & 95.84(1.49) & 95.87(1.32) & \textbf{95.98(1.37)} \\
         Vehicle & 74.17(1.83) & 74.32(2.17) &  74.53(2.1) & \textbf{75.05(2.41)} \\
       Vertebral & 82.88(3.83) & \textbf{84.36(4.28)} & 82.44(3.54) & 84.04(3.23) \\
          Voice3 & \textbf{77.58(3.35)} & 77.33(3.39) & 77.17(3.42) & 76.58(3.09) \\
         Weaning & 80.72(4.77) & 82.24(4.91) & 81.25(4.94) & \textbf{82.43(4.39)}\\
            Wine & 97.67(1.79) & 97.33(2.06) & 97.11(2.23) &  \textbf{98.0(1.71)} \\ \hline
         Average & 81.13 &  81.6 & 81.43 & \textbf{81.89} \\
         Ave Rank & 2.93& 2.26& 2.80& \textbf{1.95} \\ 
\hline
\end{tabular}}
 
	\end{table}
	
	We can see in this table that FH-DES-M achieved the highest accuracy in 13 out of 30 datasets. Additionally, this approach achieved the highest average accuracy and the best average rank in 30 datasets. In addition, using the Gabrys membership function, FH-GM proposed framework achieved the highest accuracy in 11 of 30 datasets.\par
\begin{figure}
		\centering
		\includegraphics[width=0.99\linewidth]{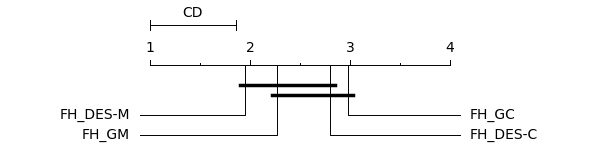}
		\caption{Critical Difference (CD) diagram for different configurations of the proposed framework. The best algorithm is the one presenting the lowest rank. Techniques that are statistically equivalent are connected by a black bar.}
		\label{fig:cddiagram_REC}
	\end{figure}

    
    In the next step, statistical analysis is conducted using the post-hoc Bonferroni-Dunn test \cite{demvsar2006statistical}. This test is applied to compare the ranks achieved by each DS method. The average ranks of different configurations of the proposed framework and the result of the Bonferroni-Dunn post-hoc test are presented in Figure \ref{fig:cddiagram_REC} using the Critical Difference (CD) diagram. The performance of the two DS approaches is significantly different if their difference in average rank is higher than the CD value.
    
	According to the post-hoc test, FH-DES-M is significantly better than FH-GC. And there is no significant difference between the other configurations of the proposed framework. However, FH-DES-M is slightly better than others. \par 
	In the next step, a pairwise comparison is conducted to compare the obtained results of FH-DES-M and FH-GM against the other configurations, based on Sign test \cite{demvsar2006statistical}. In this test, there are two hypotheses that include $H_0$ (null hypothesis) and $H_1$ (alternate hypothesis). Rejection of $H_0$ means that the performance of the corresponding DS technique is significantly better than the compared technique. The number of wins, ties, and losses for each technique is computed compared to the baseline. The significance level of this test is determined by the predefined parameter $\alpha$, which in this paper is set $\alpha = 0.05$ to have 95\% confidence. If the number of wins is greater than or equal to a critical value, denoted by $n_c$, the null hypothesis $H_0$ is rejected. The critical value is computed using equation \ref{eq:nc}:
	
	\begin{equation}
    	n_c= \frac{n_{exp}}{2} + z_\alpha \frac{\sqrt{n_{exp}}}{2}
    \label{eq:nc}
	\end{equation}
	
\begin{figure}
		\centering  
		\includegraphics[width=0.7\linewidth]{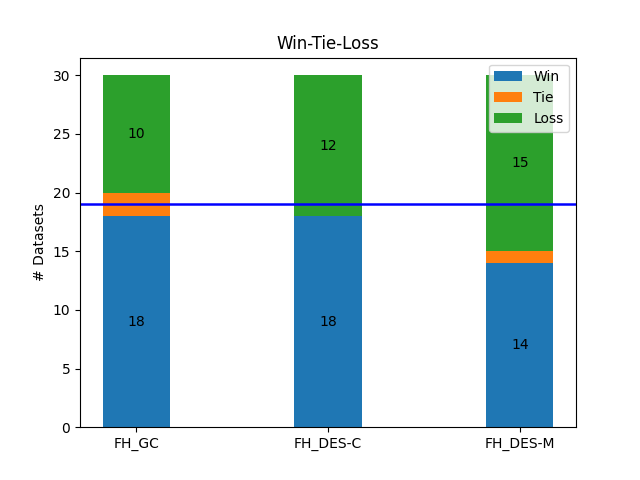}
		\caption{Pairwise comparison between the FH-GM and other configurations of the proposed approach ($n_c = 19.5$)}
		\label{fig:win-lossdiagram_FH_GM_REC}
	\end{figure}

\begin{figure}
		\centering
		\includegraphics[width=0.7\linewidth]{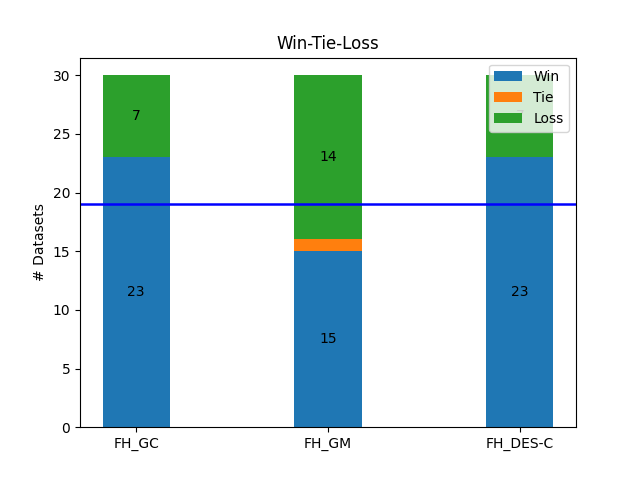}
		\caption{Pairwise comparison between the FH-DES-M and other configurations of the proposed approach ($n_c = 19.5$)}
		\label{fig:win-lossdiagram-FHDES-M_REC}
	\end{figure}

	Where $n_{exp}$ is the total number of experiments and $z_\alpha=1.645$, for a significance level of $\alpha = 0.05$ \cite{cruz_analyzing_2017}. In this test, we have 30 experiments (datasets). Therefore, $n_{exp} = 30$, so for this amount of experiments, the critical value is $n_c = 19.5$.
	Obtained results for FH-GM and FH-DES-M are represented in Figure \ref{fig:win-lossdiagram_FH_GM_REC} and Figure \ref{fig:win-lossdiagram-FHDES-M_REC} respectively.   

	It can be observed that the proposed framework statistically has good performance using both the Gabrys membership function and the proposed membership function. FH-DES-M is slightly more accurate than FH-GM, but there is no significant difference. However, these two approaches are significantly better than the correct-classified versions. Therefore, it can be concluded that the misclassified samples contain more helpful information to estimate the competence of classifiers rather than correct-classified ones. The reason is simple; boundary regions are usually challenging in classification applications. Most classification errors occur in these areas, and misclassified samples are usually found there. When we form hyperboxes based on misclassified samples, most of the hyperboxes are formed in these areas. The proximity of hyperboxes to the boundary regions means that the system makes a more precise decision in this region. Thus, it results in higher accuracy during the generalization phase. Furthermore, according to Figures \ref{fig:win-lossdiagram-FHDES-M_REC} and \ref{fig:cddiagram_REC}, FH-DES-M statistically outperformed FH-GC and FH-DES-C and was slightly better than FH-GM. Therefore, FH-DES-M was selected as the best configuration of the proposed framework and compared with other DS approaches in the next step.

	\subsection{Comparison with state-of-the-art DS methods}
	
	The accuracy and standard deviation of the proposed framework and the state-of-the-art DS approaches are reported in Table \ref{tbl:allresults}. In this table, the Oracle approach \cite{kuncheva2002theoretical} is a conceptual method that selects the base classifier which labels the query sample correctly if such base classifier exists.   
	
	\begin{table*}[t]
		\centering
		\caption{Average accuracy and standard deviation of the proposed method and other DES approaches}
		\label{tbl:allresults}
		\scalebox{0.9}{
			\begin{tabular}{ll|llllllll}
			\hline
			DataSets &Oracle &  KNORA-E&     RANK&     KNORA-U&     OLA&     DESKNN&     META-DES&     KNOP&     FH-DES-M \\
			\hline
		   Audit & 99.56(0.5) & 96.89(0.73) & 96.71(0.74) &  96.19(1.3) & \textbf{97.02(0.93)} & 96.76(0.99) & 96.74(1.04) & 96.24(1.36) & 96.87(0.92) \\
          Banana & 93.1(1.93) & 90.94(1.79) & 91.12(1.92) &  87.52(1.9) & \textbf{91.48(2.04)} &  88.2(1.87) & 88.48(1.88) &  86.6(1.71) &  89.5(1.68) \\
        Banknote & 99.87(0.22) & 99.43(0.53) & 99.43(0.69) & 98.82(0.56) & \textbf{99.45(0.68)} &  99.2(0.47) &  99.2(0.54) & 98.85(0.55) &  99.34(0.5) \\
           Blood & 89.6(2.42) & 76.74(2.24) & 76.66(2.07) &  78.24(1.6) & 77.62(2.46) &  77.67(1.9) & 77.99(1.51) & \textbf{78.37(1.55)} & 76.55(2.64) \\
          Breast & 98.99(0.56) & 96.61(1.67) & 96.15(1.88) & 96.78(1.49) & 96.12(1.76) & 96.71(1.35) & \textbf{96.82(1.44)} & 96.75(1.37) & 96.61(1.61) \\
             Car &  83.52(1.06) & 73.85(1.24) & 73.62(1.27) &  71.25(1.1) & \textbf{74.48(1.04)} & 72.31(0.79) & 73.76(0.91) & 71.27(1.07) & 74.11(1.25) \\
Datausermodeling & 99.06(1.19) & \textbf{91.44(2.99)} & 89.21(3.82) & 88.22(3.15) & 89.55(3.32) & 90.35(2.75) & 90.64(2.77) & 88.47(3.15) & 91.29(3.63) \\
          Faults & 91.54(1.21) & 69.97(1.54) & 69.12(2.17) &  69.6(2.13) & 69.42(2.16) & \textbf{70.51(2.01)} &  70.0(2.08) &  69.7(2.15) & 70.38(2.03) \\
          German & 94.54(1.45) & 74.46(2.75) & 73.86(2.35) & 75.08(1.87) &  73.92(2.9) & \textbf{75.22(1.96)} & 74.94(1.93) &  75.04(2.0) & 74.86(2.27) \\
        Haberman & 92.4(2.06) & 70.71(4.32) &  70.84(4.2) &  73.9(3.15) & 71.36(4.24) &  74.03(2.6) &  72.21(3.3) & \textbf{74.16(2.75)} & 71.56(3.98) \\
        
           Heart &  97.35(2.01) & 82.35(4.26) & 81.47(4.42) &  \textbf{83.9(3.93)} & 82.65(4.76) & 83.75(3.96) & 83.24(3.93) &  83.75(3.9) &  \textbf{83.9(4.42)} \\
            ILPD & 95.68(1.62) & 70.31(3.04) & 69.97(2.88) & \textbf{72.09(2.08)} & 70.75(3.07) & 70.89(1.57) &  71.3(2.48) & 71.64(1.72) & 70.65(2.59) \\
      Ionosphere &  98.18(1.32) &\textbf{ 89.32(1.66)} & 86.53(2.39) & 88.07(2.02) & 86.36(2.49) & 88.24(2.25) & 88.58(2.45) & 87.95(2.47) & 88.13(1.37) \\
      Laryngeal1 &  95.28(3.82) & 82.41(3.68) & 82.13(4.15) & 82.41(4.08) & 82.04(4.02) &   82.5(4.2) & 82.13(4.73) & \textbf{82.59(4.07)} &  82.5(3.49) \\
      Laryngeal3 & 89.38(2.65) & 71.18(3.26) & 71.24(4.73) & 71.29(3.73) &  71.29(4.8) &  71.4(3.67) &  70.9(3.59) & 71.18(3.49) & \textbf{71.8(4.08)} \\
      Lithuanian &  93.47(1.78) &   91.1(2.1) &  \textbf{91.1(2.45)} &   86.9(2.0) &  90.9(2.31) &  88.6(2.34) & 87.93(1.87) & 86.37(1.89) &  90.5(2.32) \\
           Liver & 97.93(1.91) & 68.16(4.45) & 67.64(3.26) & 68.91(4.58) & 69.66(3.96) & \textbf{70.86(4.23)} & 68.74(3.94) &  68.85(4.6) & 69.14(4.34) \\
    Mammographic &  90.29(2.25) & 77.81(2.77) & 77.55(2.94) &  79.04(2.5) & 78.51(2.14) & \textbf{79.42(2.45)} & 78.49(2.38) & 79.13(2.55) & 78.87(2.73) \\
           Monk2 & 97.31(1.01)  & 86.99(3.46) & 85.69(3.33) & 79.44(3.63) & 84.81(3.01) & 83.06(3.53) & \textbf{88.75(3.99)} &  81.3(3.69) & 87.64(3.24) \\
         Phoneme &  87.75(1.75) & \textbf{79.81(1.01)} & 79.62(1.03) & 77.49(0.92) & 79.09(1.03) & 77.77(0.81) & 79.51(1.05) & 77.28(0.89) &  78.1(0.91) \\
            Pima & 92.97(1.79) & 76.12(2.53) & 75.42(2.72) & 77.03(2.08) & 75.49(2.51) & 76.54(2.45) & 76.59(2.47) & \textbf{77.21(2.08)} & 76.28(2.72) \\
           Sonar & 98.85(1.54) & 78.08(5.21) & 77.02(4.61) & 77.31(5.69) & 76.63(5.17) & 77.98(5.49) &  \textbf{80.0(5.28)} &  77.4(5.83) & 79.62(5.42) \\
         Statlog & 94.14(1.53) & 74.72(2.42) & 74.48(1.97) & 75.22(2.33) &  74.64(2.2) & 75.42(2.08) &  \textbf{75.82(2.1)} & 75.42(2.43) &  75.08(2.0) \\
           Steel & 91.55(1.44) & 70.67(1.45) &  69.9(1.94) &  70.3(1.81) & 70.02(1.68) & \textbf{71.65(1.81)} & 71.05(1.77) & 70.64(1.92) & 71.37(1.33) \\
         Thyroid &  98.47(0.88) & 95.84(1.43) &  95.9(1.34) &  95.9(1.37) & 95.78(1.17) &  95.78(1.2) & 95.95(1.28) & 95.84(1.34) & \textbf{95.98(1.37)} \\
         Vehicle & 96.89(1.02) & 74.98(2.17) &  74.83(2.5) & 74.79(1.67) & 74.25(2.07) & 74.55(2.47) & 74.81(2.24) &  74.5(1.82) & \textbf{75.05(2.41)} \\
       Vertebral & 95.96(2.93) & 83.33(4.05) & 83.08(3.88) & 82.44(4.27) & 83.27(3.35) &  \textbf{84.1(4.82)} & 83.33(3.85) & 83.01(4.42) & 84.04(3.23) \\
          Voice3 &  92.67(2.32) &  77.0(3.32) & 76.92(3.62) & 78.58(2.85) & 77.08(3.61) & 77.83(2.79) & 77.83(3.42) & \textbf{78.75(2.58)} & 76.58(3.09) \\
         Weaning &  97.43(2.06) & 81.32(4.26) & 80.92(4.51) &  80.99(4.6) &  81.91(4.9) &  \textbf{82.5(4.25)} & 81.18(4.37) & 80.92(4.66) & 82.43(4.39) \\
            Wine & 99.78(0.67) & 97.89(1.79) & 96.56(3.02) & 97.78(1.86) & 96.67(2.77) & \textbf{98.22(1.66)} & 97.78(2.11) & 97.89(1.92) &  98.0(1.71) \\ \hline
         
         Average & 94.78 &  81.68 & 81.16 & 81.18 & 81.41 & 81.73 & 81.82 & 81.24 & \textbf{81.89} \\
	Ave Rank & - & 3.32  & 5.1 & 3.88 & 4.18 & \textbf{2.42} & 2.85 & 3.78 & 2.47 \\  \hline
		\end{tabular} }
	\end{table*}
	We can see in Table \ref{tbl:allresults} that the proposed approach FH-DES-M achieved the highest average accuracy among all DS approaches, and its average rank is very close to the DESKNN that obtained the lowest rank.\par
\begin{figure}
		\centering
		\includegraphics[width=0.99\linewidth]{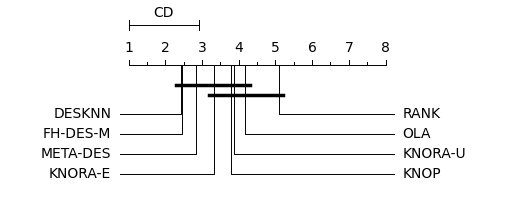}
		\caption{Critical Difference (CD) diagram considering the all compared approaches. The best algorithm is the one presenting the lowest rank and techniques that are statistically equivalent are connected by a black bar.}
		\label{fig:cddiagram2}
	\end{figure}

 
	Figure \ref{fig:cddiagram2} presents the average ranks of different DS techniques and the result of the Bonferroni-Dunn post-hoc test using a critical difference diagram.
	
\begin{figure}
		\centering
		\includegraphics[width=0.9\linewidth]{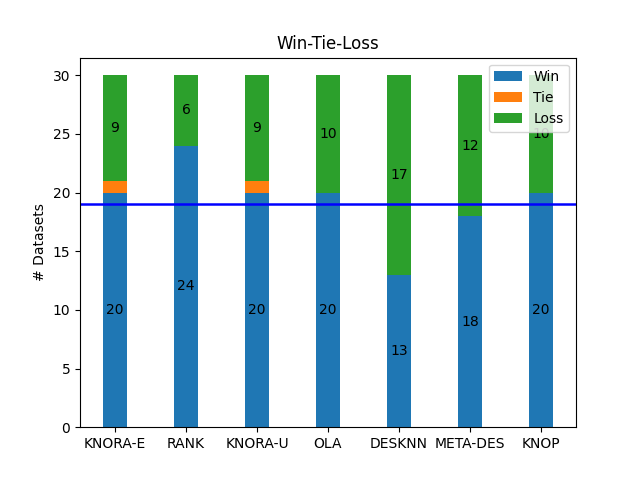}
		\caption{Pairwise comparison between the FH-DES-M and other DS methods ($n_c = 19.5$)}
		\label{fig:win-lossdiagram-FHDES-M}
	\end{figure}

	This figure shows that DESKNN and FH-DES-M have the best overall rank among the compared methods, outperforming all DS approaches. However, there is no significant difference between these DES approaches over the mentioned datasets. For a more fine-grained comparison between these techniques, we conducted a pairwise comparison between FH-DES-M and the state-of-the-art DS techniques using the Sign test. The result of this test is shown in Figure \ref{fig:win-lossdiagram-FHDES-M}.
	
 As we can see in this figure, the proposed approach significantly surpasses five out of seven state-of-the-art DS approaches. And it is slightly better than META-DES. Just DESKNN approach has a higher number of wins than FH-DES-M (17 losses for the FH-DES-M). But, since the difference is lower than the threshold $n_c = 19.5$, this difference is not statistically significant.  

	\subsection{Time complexity and Memory Cost}
	\label{subse:Time-Memory}
	As we discussed earlier, in the KNN-based approaches, to label each query sample, its distance to all DSEL samples is calculated, and then the K nearest samples are determined. If we used big \textit{O} notation that represents an upper bound to show how run time grow as the input size grows \cite{knuth1976big}, KNN costs \textit{$O(n)$}, which \textit{n} is the number of instances in DSEL.
	However, the computational complexity in the proposed approach changes to $O(e)$ which \textit{e} is the number of hyperboxes. We expect that the number of generated hyperboxes is much smaller than the size of the DSEL data ($e << n$), significantly reducing the computational cost of our system. 
	To validate this hypothesis, we analyze the number of hyperboxes generated by our system considering two large scale datasets:
	i) the \textit{Sensor} dataset \cite{asuncion2007uci} which contains 928K samples as a binary classification problem. ii) An artificial dataset, namely \textit{ArData} generated with Scikit-learn, which includes five features and two classes. We vary the DSEL size from 1K to 900K to examine the influence of its size on the number of generated hyperboxes. In each step, we added new samples to the DSEL data. In this experiment, two small subsets of the dataset are selected randomly as train and test data so that each of them contains 1000 samples. During all steps of this experiment, there was no overlap between the test, train, and DSEL datasets. The pool of classifiers contained 100 perceptrons, similar to the previous experiments. \par  
	Since the complexity of all KNN based approaches is the same ($O(n)$), during this experiment, we examined only the META-DES approach as a strong and accurate KNN-based approach \cite{cruz_dynamic_2018}. In Table \ref{tbl:largeScale_Accuracy}, the accuracy obtained using FH-DES-M and META-DES in different sizes of the DSEL data is reported. 
	
\begin{table}
\centering
\caption{The accuracy of the proposed approach and META-DES in different data sizes}
\label{tbl:largeScale_Accuracy}
\scalebox{1.0}{
    \begin{tabular}{llll} 
    \hline
         DataSets &      Oracle &         META-DES &    FH-DES-M  \\
\hline
    Data1000 & 94.01(0.19) & 90.78(0.07) & 90.99(0.11) \\
   Data10000 & 94.01(0.19) & 91.06(0.14) & 90.74(0.13) \\
  Data100000 & 94.01(0.19) & 91.74(0.09) & 90.84(0.19) \\
  Data300000 & 94.01(0.19) & 91.99(0.08) & 90.84(0.17) \\
  Data500000 &  94.01(0.19) & 92.07(0.15) & 90.82(0.09) \\
  Data700000 &  94.01(0.19) & 92.06(0.17) & 90.80(0.10) \\ 
  Data900000 &  94.01(0.19) & 92.06(0.16) & 90.86(0.14) \\ \hline
  Sensor1000 & 99.09(0.14) & 97.68(0.2) &  97.26(0.4) \\
 Sensor10000 & 99.09(0.14) & 98.21(0.19) &  97.4(0.17) \\
Sensor100000 & 99.09(0.14) & 98.75(0.16) & 96.61(0.38) \\
Sensor300000 & 99.09(0.14) & 98.78(0.16) & 96.39(0.25) \\
Sensor500000 & 99.09(0.14) & 98.84(0.15) & 96.36(0.23) \\
Sensor700000 & 99.09(0.14) & 98.84(0.16) & 96.36(0.28) \\
Sensor900000 & 99.09(0.14) & 98.87(0.15) & 96.38(0.29) \\ \hline
Average Accuracy & 96.74 & 95.37 & 93.98 \\ \hline
    \end{tabular}}
    
\end{table}	

This table shows that the proposed method and META-DES have similar accuracy in small data sizes. However, with larger data size, the accuracy of the KNN-based approach increases. In particular, the accuracy of the META-DES gets close to the oracle's accuracy for the Sensor dataset as the dataset size increases. Because, in large-scale datasets, almost all regions of the feature space are covered by DSEL instances. Thus, sufficient DSEL samples are available to correctly estimate the base classifiers' competencies and select a suitable ensemble of classifiers. However, achieving such accuracy has a large computational complexity. In table \ref{tbl:largescale}, the number of hyperboxes generated by FH-DES-M and the number of DSEL samples are reported. 
	
\begin{table}[]
    \centering
    \caption{Comparing the number of generated Hyperboxes in FH-DES-M and number of DSEL samples}
    \label{tbl:largescale}
    \begin{tabular}{l|ll}
    \hline
    \multirow{2}{*}{\#Sample} & \multicolumn{2}{c}{\# Hyperboxes} \\
                  & ArData & Sensor \\ \hline
        1,000    &  530   &  1,115 \\
        10,000   &  1,313 &   2,407\\
        100,000  &  2,679 &   3,849\\
        300,000  &  3,457 &   4,167\\
        500,000  &  3,818 &   4,290\\
        700,000  &  4,026 &   4,331\\
        900,000  &  4,127 &   4,392\\ \hline
         
    \end{tabular}
    
\end{table}
As presented in table~\ref{tbl:largescale}, the number of generated hyperboxes is considerably smaller than the size of the DSEL data ($e << n$). For example, considering the 900K samples study case, the proposed approach generates only 4,127 and 4,392  hyperboxes for the ArData and Sensor datasets, respectively. In addition, we can also observe a plateau in the number of hyperboxes added to the system as the dataset size increases. Between 700k to 900k samples, only 101 and 61 new hyperboxes were added to the system. On the other hand, all 900K samples should be stored and processed for each query sample using KNN-based approaches. Thus, we can confirm that the proposed approach has lower computational complexity than KNN-based DES approaches from the storage and computational perspectives.


\section{Conclusion}
\label{sec:conclusion}
	This paper introduced a novel dynamic ensemble selection framework based on fuzzy hyperbox. In this framework, competence or incompetence areas of classifiers are determined using fuzzy hyperboxes. For each base classifier, hyperboxes are formed based on their correct-classified samples to define the "competencies" and "incompetencies" areas. In addition, for the first time in the dynamic selection area, the misclassified instances were applied to define the incompetence areas. Moreover, this paper also introduced a new membership function to measure the memberships differently with softer boundaries that slightly increase the system's accuracy. \par 
	
	Experimental results demonstrated that utilizing the misclassified samples could significantly increase the accuracy of the proposed DS framework and decrease the computational complexity compared to the correct classified samples. Additionally, the proposed approach based on misclassified samples obtained the highest average accuracy compared to state-of-the-art DS approaches.\par 
	
	Furthermore, the proposed framework also has lower storage and computational complexity when compared to DS techniques based on KNN and potential function models for estimating the regions of competence. According to the experimental results, the proposed method generates about 4k hyperboxes (40 hyperboxes for modeling each base classifier's misclassifications) in datasets containing 900k samples. Thus, the proposed FH-DES can be an excellent alternative for handling large-scale problems with DES approaches. Fuzzy hyperboxes also allow online learning making it a suitable technique for handling streaming data. Future works will investigate the use of FH-DES for dealing with data streams and concept drift.
	






\bibliographystyle{IEEEtran}
%

\bibliography{IJCNN2022.bib}

\begin{thebibliography}{10}
\providecommand{\url}[1]{#1}
\csname url@samestyle\endcsname
\providecommand{\newblock}{\relax}
\providecommand{\bibinfo}[2]{#2}
\providecommand{\BIBentrySTDinterwordspacing}{\spaceskip=0pt\relax}
\providecommand{\BIBentryALTinterwordstretchfactor}{4}
\providecommand{\BIBentryALTinterwordspacing}{\spaceskip=\fontdimen2\font plus
\BIBentryALTinterwordstretchfactor\fontdimen3\font minus
  \fontdimen4\font\relax}
\providecommand{\BIBforeignlanguage}[2]{{%
\expandafter\ifx\csname l@#1\endcsname\relax
\typeout{** WARNING: IEEEtran.bst: No hyphenation pattern has been}%
\typeout{** loaded for the language `#1'. Using the pattern for}%
\typeout{** the default language instead.}%
\else
\language=\csname l@#1\endcsname
\fi
#2}}
\providecommand{\BIBdecl}{\relax}
\BIBdecl

\bibitem{cruz_dynamic_2018}
\BIBentryALTinterwordspacing
R.~M. Cruz, R.~Sabourin, and G.~D. Cavalcanti,
  ``\BIBforeignlanguage{en}{Dynamic classifier selection: {Recent} advances and
  perspectives},'' \emph{\BIBforeignlanguage{en}{Information Fusion}}, vol.~41,
  pp. 195--216, May 2018. [Online]. Available:
  \url{https://linkinghub.elsevier.com/retrieve/pii/S1566253517304074}
\BIBentrySTDinterwordspacing

\bibitem{zyblewski2021preprocessed}
P.~Zyblewski, R.~Sabourin, and M.~Wo{\'z}niak, ``Preprocessed dynamic
  classifier ensemble selection for highly imbalanced drifted data streams,''
  \emph{Information Fusion}, vol.~66, pp. 138--154, 2021.

\bibitem{britto2014dynamic}
A.~S. Britto~Jr, R.~Sabourin, and L.~E. Oliveira, ``Dynamic selection of
  classifiers—a comprehensive review,'' \emph{Pattern recognition}, vol.~47,
  no.~11, pp. 3665--3680, 2014.

\bibitem{kuncheva2014combining}
L.~I. Kuncheva, \emph{Combining pattern classifiers: methods and
  algorithms}.\hskip 1em plus 0.5em minus 0.4em\relax John Wiley \& Sons, 2014.

\bibitem{cruz_meta-des__2015}
R.~M.~O. Cruz, ``\BIBforeignlanguage{en}{{META}-{DES}\_ {A} dynamic ensemble
  selection framework using meta-learning},''
  \emph{\BIBforeignlanguage{en}{Pattern Recognition}}, p.~11, 2015.

\bibitem{fernandez-delgado_we_nodate_2014}
M.~Fern{\'a}ndez-Delgado, E.~Cernadas, S.~Barro, and D.~Amorim,
  ``\BIBforeignlanguage{en}{Do we need hundreds of classifiers to solve real
  world classification problems?}'' \emph{\BIBforeignlanguage{en}{The journal
  of machine learning research}}, vol.~15, no.~1, pp. 3133--3181, 2014.

\bibitem{xiao_ensemble_2016}
\BIBentryALTinterwordspacing
H.~Xiao, Z.~Xiao, and Y.~Wang, ``\BIBforeignlanguage{en}{Ensemble
  classification based on supervised clustering for credit scoring},''
  \emph{\BIBforeignlanguage{en}{Applied Soft Computing}}, vol.~43, pp. 73--86,
  Jun. 2016. [Online]. Available:
  \url{https://linkinghub.elsevier.com/retrieve/pii/S1568494616300734}
\BIBentrySTDinterwordspacing

\bibitem{krawczyk_dynamic_2018}
B.~Krawczyk, ``\BIBforeignlanguage{en}{Dynamic ensemble selection for
  multi-class classification with one-class classifiers},''
  \emph{\BIBforeignlanguage{en}{Pattern Recognition}}, p.~18, 2018.

\bibitem{elmi_2020_fuzzyHesitate}
J.~Elmi, ``\BIBforeignlanguage{en}{Dynamic ensemble selection based on hesitant
  fuzzy multiple criteria decision making},'' p.~13.

\bibitem{kuncheva_clustering-and-selectionmodel_nodate}
L.~I. Kuncheva, ``\BIBforeignlanguage{en}{Clustering-and-{SelectionModel} for
  {Classifier} {Combinat} ion},'' p.~4.

\bibitem{lin_libd3c_2014}
\BIBentryALTinterwordspacing
C.~Lin, W.~Chen, C.~Qiu, Y.~Wu, S.~Krishnan, and Q.~Zou,
  ``\BIBforeignlanguage{en}{{LibD3C}: {Ensemble} classifiers with a clustering
  and dynamic selection strategy},''
  \emph{\BIBforeignlanguage{en}{Neurocomputing}}, vol. 123, pp. 424--435, Jan.
  2014. [Online]. Available:
  \url{https://linkinghub.elsevier.com/retrieve/pii/S0925231213007959}
\BIBentrySTDinterwordspacing

\bibitem{woloszynski_new_2009}
T.~Woloszynski and M.~Kurzynski, ``On a {New} {Measure} of {Classifier}
  {Competence} {Applied} to the {Design} of {Multiclassifier} {Systems},'' in
  \emph{Image {Analysis} and {Processing} – {ICIAP} 2009}, P.~Foggia,
  C.~Sansone, and M.~Vento, Eds.\hskip 1em plus 0.5em minus 0.4em\relax Berlin,
  Heidelberg: Springer Berlin Heidelberg, 2009, pp. 995--1004.

\bibitem{woloszynski_measure_2012}
\BIBentryALTinterwordspacing
T.~Woloszynski, M.~Kurzynski, P.~Podsiadlo, and G.~W. Stachowiak,
  ``\BIBforeignlanguage{en}{A measure of competence based on random
  classification for dynamic ensemble selection},''
  \emph{\BIBforeignlanguage{en}{Information Fusion}}, vol.~13, no.~3, pp.
  207--213, Jul. 2012. [Online]. Available:
  \url{https://linkinghub.elsevier.com/retrieve/pii/S1566253511000297}
\BIBentrySTDinterwordspacing

\bibitem{giacinto_dynamic_2001}
G.~Giacinto and F.~Roli, ``\BIBforeignlanguage{en}{Dynamic classi"er selection
  based on multiple classi"er behaviour},''
  \emph{\BIBforeignlanguage{en}{Pattern Recognition}}, p.~3, 2001.

\bibitem{cavalin_logid_2012}
P.~R. Cavalin, ``\BIBforeignlanguage{en}{{LoGID} {An} adaptive framework
  combining local and global incremental learning for dynamic selection of
  ensembles of {HMMs}},'' \emph{\BIBforeignlanguage{en}{Pattern Recognition}},
  p.~13, 2012.

\bibitem{batista_dynamic_2012}
\BIBentryALTinterwordspacing
L.~Batista, E.~Granger, and R.~Sabourin, ``\BIBforeignlanguage{en}{Dynamic
  selection of generative–discriminative ensembles for off-line signature
  verification},'' \emph{\BIBforeignlanguage{en}{Pattern Recognition}},
  vol.~45, no.~4, pp. 1326--1340, Apr. 2012. [Online]. Available:
  \url{https://linkinghub.elsevier.com/retrieve/pii/S0031320311004353}
\BIBentrySTDinterwordspacing

\bibitem{nguyen_ensemble_2020}
\BIBentryALTinterwordspacing
T.~T. Nguyen, A.~V. Luong, M.~T. Dang, A.~W.-C. Liew, and J.~McCall,
  ``\BIBforeignlanguage{en}{Ensemble {Selection} based on {Classifier}
  {Prediction} {Confidence}},'' \emph{\BIBforeignlanguage{en}{Pattern
  Recognition}}, vol. 100, p. 107104, Apr. 2020. [Online]. Available:
  \url{https://linkinghub.elsevier.com/retrieve/pii/S0031320319304054}
\BIBentrySTDinterwordspacing

\bibitem{simpson1992Classification}
P.~K. Simpson, ``Fuzzy min—max neural networks—part 1: Classification,''
  \emph{IEEE Trans. on Neural Networks}, vol.~3, no.~5, pp. 776--786, 1992.

\bibitem{simpson_fuzzy_1993}
\BIBentryALTinterwordspacing
P.~Simpson and G.~Jahns, ``\BIBforeignlanguage{en}{Fuzzy min-max neural
  networks for function approximation},'' in
  \emph{\BIBforeignlanguage{en}{{IEEE} {International} {Conference} on {Neural}
  {Networks}}}.\hskip 1em plus 0.5em minus 0.4em\relax San Francisco, CA, USA:
  IEEE, 1993, pp. 1967--1972. [Online]. Available:
  \url{http://ieeexplore.ieee.org/document/298858/}
\BIBentrySTDinterwordspacing

\bibitem{khuat_hyperbox_2019}
\BIBentryALTinterwordspacing
T.~T. Khuat, D.~Ruta, and B.~Gabrys, ``\BIBforeignlanguage{en}{Hyperbox based
  machine learning algorithms: {A} comprehensive survey},''
  \emph{\BIBforeignlanguage{en}{arXiv:1901.11303 [cs, stat]}}, Mar. 2019,
  arXiv: 1901.11303. [Online]. Available: \url{http://arxiv.org/abs/1901.11303}
\BIBentrySTDinterwordspacing

\bibitem{gabrys_general_2000}
\BIBentryALTinterwordspacing
B.~Gabrys and A.~Bargiela, ``\BIBforeignlanguage{en}{General fuzzy min-max
  neural network for clustering and classification},''
  \emph{\BIBforeignlanguage{en}{IEEE Transactions on Neural Networks}},
  vol.~11, no.~3, pp. 769--783, May 2000. [Online]. Available:
  \url{http://ieeexplore.ieee.org/document/846747/}
\BIBentrySTDinterwordspacing

\bibitem{khuat2020hyperbox}
T.~T. Khuat, D.~Ruta, and B.~Gabrys, ``Hyperbox-based machine learning
  algorithms: a comprehensive survey,'' \emph{Soft Computing}, pp. 1--39, 2020.

\bibitem{breiman1996bagging}
L.~Breiman, ``Bagging predictors,'' \emph{Machine learning}, vol.~24, no.~2,
  pp. 123--140, 1996.

\bibitem{cruz_deslib_nodate}
R.~M. Cruz, L.~G. Hafemann, R.~Sabourin, and G.~D. Cavalcanti, ``Deslib: A
  dynamic ensemble selection library in python.'' \emph{J. Mach. Learn. Res.},
  vol.~21, no.~8, pp. 1--5, 2020.

\bibitem{openml-van2013}
J.~N. Van~Rijn, B.~Bischl, L.~Torgo, B.~Gao, V.~Umaashankar, S.~Fischer,
  P.~Winter, B.~Wiswedel, M.~R. Berthold, and J.~Vanschoren, ``Openml: A
  collaborative science platform,'' in \emph{Joint european conference on
  machine learning and knowledge discovery in databases}.\hskip 1em plus 0.5em
  minus 0.4em\relax Springer, 2013, pp. 645--649.

\bibitem{asuncion2007uci}
A.~Asuncion and D.~Newman, ``Uci machine learning repository,'' 2007.

\bibitem{demvsar2006statistical}
J.~Dem{\v{s}}ar, ``Statistical comparisons of classifiers over multiple data
  sets,'' \emph{The Journal of Machine Learning Research}, vol.~7, pp. 1--30,
  2006.

\bibitem{cruz_analyzing_2017}
\BIBentryALTinterwordspacing
R.~M.~O. Cruz, R.~Sabourin, and G.~D.~C. Cavalcanti,
  ``\BIBforeignlanguage{en}{Analyzing different prototype selection techniques
  for dynamic classifier and ensemble selection},'' in
  \emph{\BIBforeignlanguage{en}{2017 {International} {Joint} {Conference} on
  {Neural} {Networks} ({IJCNN})}}.\hskip 1em plus 0.5em minus 0.4em\relax
  Anchorage, AK, USA: IEEE, May 2017, pp. 3959--3966. [Online]. Available:
  \url{http://ieeexplore.ieee.org/document/7966355/}
\BIBentrySTDinterwordspacing

\bibitem{kuncheva2002theoretical}
L.~I. Kuncheva, ``A theoretical study on six classifier fusion strategies,''
  \emph{IEEE Transactions on pattern analysis and machine intelligence},
  vol.~24, no.~2, pp. 281--286, 2002.

\bibitem{knuth1976big}
D.~E. Knuth, ``Big omicron and big omega and big theta,'' \emph{ACM Sigact
  News}, vol.~8, no.~2, pp. 18--24, 1976.

\end{thebibliography}


\end{document}